\documentclass[letterpaper]{article} 
\usepackage{aaai24}  
\usepackage{times}  
\usepackage{helvet}  
\usepackage{courier}  
\usepackage[hyphens]{url}  
\usepackage{graphicx} 
\usepackage{natbib}  
\usepackage{caption} 
\usepackage{algorithm}
\usepackage{algorithmic}
\usepackage{bm}
\usepackage{newfloat}
\usepackage{listings}
\DeclareCaptionStyle{ruled}{labelfont=normalfont,labelsep=colon,strut=off} 
\floatstyle{ruled}
\newfloat{listing}{tb}{lst}{}
\floatname{listing}{Listing}
\usepackage{amsmath}
\usepackage{amsthm}
\usepackage{amsfonts}
\usepackage[table,xcdraw]{xcolor}
\definecolor{light-gray}{gray}{0.8}
\usepackage{pgfplots}
\usepackage{multirow}
\usepackage{comment}

\newtheorem{example}{Example}
\newtheorem{predef}{Definition}
\newenvironment{definition}{\begin{predef}
\begin{rm}}{\end{rm}
\end{predef}}
\newtheorem{proposition}{Proposition}
\newtheorem{conjecture}{Conjecture}

\usepackage{xifthen}
\newboolean{remove}

\usepackage{comment}

\setboolean{remove}{false}

\ifremove
  \excludecomment{remove}
\else
  
\fi

\usepackage{tikz,array,accents}
\usetikzlibrary{shapes,shadows,calc,arrows,fit,positioning,quotes,backgrounds,chains,arrows.meta,intersections}
\usetikzlibrary{calc, chains, decorations.pathmorphing}

\tikzset{
    position/.style args={#1:#2 from #3}{
        at=(#3.#1), anchor=#1+180, shift=(#1:#2)
    },
    arg/.style={align=center,draw,circle,fill=white,minimum size=0.6cm,inner sep=-1pt},
    targ/.style={align=center,draw,circle,fill=black!10,minimum size=0.5cm,inner sep=0cm}}

\tikzset{
    position/.style args={#1:#2 from #3}{
        at=(#3.#1), anchor=#1+180, shift=(#1:#2)
    },
    arg/.style={align=center,draw,circle,fill=white,minimum size=0.6cm,inner sep=-1pt},
    barg/.style={align=center,draw,circle,fill=white,minimum size=1cm,inner sep=-1pt},
    marg/.style={align=center,draw,circle,fill=white,minimum size=1cm,inner sep=-1pt,font={\scriptsize}},
    }

\ifx\text\undefined
  \newcommand{\text}[1]{\relax
    \ifmmode\mathchoice
      {\hbox{\the\textfont0\relax#1}}%
      {\hbox{\the\textfont0\relax#1}}%
      {\hbox{\the\scriptfont0\relax#1}}%
      {\hbox{\the\scriptscriptfont0\relax#1}}%
    \else{\relax#1}\fi}
  \fi
\usetikzlibrary{decorations.markings}
\tikzset{
  negate/.style={
    decoration={
      markings,
      mark= at position 0.5 with {
        \node[transform shape] (tempnode) {$/$};
      },
    },
    postaction={decorate},
  },
}

\newcommand{\dash}[1]{{\operatorname{\mathit{#1}}}}

\title{An Argumentation-based Approach for Representing Individual Fairness}
\title{An Argumentation-Based Approach for Identifying Reasons for Bias}
\title{Identifying Reasons for Bias: An Argumentation-Based Approach}
\author{
    Madeleine Waller\textsuperscript{\rm 1}, Odinaldo Rodrigues\textsuperscript{\rm 1}, Oana Cocarascu\textsuperscript{\rm 1}
}
\affiliations{
    \textsuperscript{\rm 1}King's College London\\

    \{madeleine.waller, odinaldo.rodrigues, oana.cocarascu\}@kcl.ac.uk
}

\begin{document}

\maketitle

\begin{abstract}
As algorithmic decision-making systems become more prevalent in society, ensuring the fairness of these systems is becoming increasingly important. Whilst there has been substantial research in building fair algorithmic decision-making systems, the majority of these methods require access to the training data, including personal characteristics, and are not transparent regarding which individuals are classified unfairly. In this paper, we propose a novel model-agnostic argumentation-based method to determine why an individual is classified differently in comparison to similar individuals. Our method uses a quantitative argumentation framework to represent attribute-value pairs of an individual and of those similar to them, and uses a well-known semantics to identify the attribute-value pairs in the individual contributing most to their different classification. We evaluate our method on two datasets commonly used in the fairness literature and illustrate its effectiveness in the identification of bias.

\end{abstract}
\newcommand{\oldtext}[1]{\textcolor{red}{#1}}
\newcommand{\newtext}[1]{\textcolor{orange}{#1}}

\section{Introduction\label{sec:introduction}}

As machine learning (ML) algorithms are increasingly used in decision-making systems with high impact on individuals, there is a need to ensure not only that the decisions made are fair, but also to explain the decision to the individual affected.
A system is considered to be fair if it does not discriminate based on protected personal characteristics such as race, sex, religion, etc.
There have been instances where decision-making systems discriminated against individuals in domains such as criminal justice \cite{thepartnershiponaiReportAlgorithmicRisk2019}, recruitment \cite{DBLP:journals/ethicsit/Tilmes22}, and social services \cite{gillinghamDecisionSupportSystems2019}.
An analysis of COMPAS \cite{northpointePractitionerGuideCOMPAS2019}, a popular tool used in the US to predict whether criminals will re-offend, found that black defendants were identified incorrectly as re-offending at a higher rate than white defendants \cite{larsonHowWeAnalyzed2016}.

Research on fairness has received increased attention in recent years and several fairness metrics have been developed to quantify the fairness of a system (see~\citet{Mehrabi:21} for an overview on bias and fairness in machine learning). These metrics can be classified into group fairness (i.e. detecting bias across different values of a protected attribute, e.g., male and female individuals \cite{DBLP:conf/bigdataconf/GargVF20}) and individual fairness (i.e. detecting bias for an individual compared to similar individuals \cite{DBLP:conf/icml/MukherjeeYBS20}).
Whilst several notions of evaluating fairness have been proposed in the literature, there is no agreement as to which fairness metric to apply in which scenario \cite{Verma:18}. Furthermore, interpreting the meaning of the values returned by a metric is not always intuitive; for example,  simply reporting the percentage level of fairness of a system (e.g., 80\%) may not give full confidence for stakeholders in the system.
Most existing group metrics also require the specification of protected attributes and can only detect unwanted bias with respect to one binary protected attribute. Finally, quantifying fairness requires full access to the training data and protected attributes in order to measure the difference in positive  classifications across protected groups. In reality, this data may not be available and it may be difficult to pre-define protected attributes before deploying the system \cite{DBLP:conf/ssci/HaeriZ20}.

Identifying a link between the input data and the final decision is an important step towards providing a fair and transparent explanation \cite{Hamon:22}. Computational argumentation has long been seen as a means for explaining reasoning (see \citet{Vassiliades:P21,Cyras:21} for an overview). Specifically, abstract Argumentation Frameworks (AFs) were proposed as a way to represent and reason with conflicting information \cite{DBLP:journals/ai/Dung95}. Several types of semantics have been proposed to evaluate the acceptability of arguments in AFs \cite{DBLP:journals/argcom/BaroniRTAB15} and their extensions. AFs have been used for a variety of applications such as decision-making systems \cite{DBLP:journals/ai/AmgoudP09,DBLP:journals/expert/BrardaTG21}, recommender systems \cite{DBLP:conf/atal/CocarascuRT19,DBLP:conf/ijcai/RagoCT18}, knowledge-based systems \cite{DBLP:conf/eumas/KokciyanPSSM20}, and planning and scheduling systems \cite{DBLP:conf/aaai/CyrasLMT19}. However, until now they have not been explored in relation to individual fairness in decision-making systems.

In this paper, we propose a novel argumentation-based approach for identifying bias in relation to individual fairness which does not require access to labelled data, the training algorithm, or the specification of protected attributes before deployment. We focus on individual fairness as subjects of decisions will mostly be concerned about their personal treatment, rather than any group. Hence we move away from quantifying fairness using existing group fairness metrics and offer a transparent representation of the reasons for a classification from which an explanation can be extracted.

We use a quantitative argumentation framework to represent the arguments of similar individuals that reason why the queried individual received a classification.
Reasons are differences in the values of attributes in the queried individual in relation to those of similar individuals with different classifications. The strength of attacks between attribute-value pairs is calculated as the proportion of similar individuals with particular characteristics and the overall evaluation is done using the weighted h-Categorizer semantics \cite{DBLP:journals/ai/AmgoudDV22} which calculates the final weights of arguments. As a result, final weights correspond to the attribute-value pairs that contribute most to the negative classification of a queried individual compared to similar individuals.

\section{Background}
In this section, we provide the core background on fairness in ML and argumentation on which our method relies.

\subsection{Fairness in ML}

Fairness in ML involves ensuring decision-making ML systems are fair for individuals and for different groups defined by protected personal characteristics. Various metrics have been proposed to quantify fairness in decision-making systems \cite{Mehrabi:21}.

Individual fairness metrics determine whether similar individuals receive the same classifications \cite{DBLP:conf/innovations/DworkHPRZ12}.
For example, \citet{DBLP:conf/icml/ZemelWSPD13} counts the pairs of individuals receiving the same classification, where the notion of similarity depends on the context of the application. Group fairness metrics have been defined as the difference in the number of positive and negative classifications across two protected groups~\cite{Mehrabi:21}. For example, demographic parity is calculated as the proportion of positive classifications for the protected group divided by the proportion for the non-protected group~\cite{DBLP:conf/icdm/CaldersKP09}.

Bias detection methods are used to assess the fairness of a system, and bias mitigation methods attempt to make a system fairer with respect to some metric(s). Mitigation methods target different stages of the ML system including the pre-processing of the training data (e.g. \cite{DBLP:journals/kais/KamiranC11,DBLP:conf/icdm/ZliobaiteKC11,DBLP:conf/kdd/FeldmanFMSV15}), the training algorithm (e.g. \cite{DBLP:conf/dis/HuILZYNR20,DBLP:conf/cikm/IosifidisN19,DBLP:conf/aies/OnetoDEP19}), or in the classifications (e.g. \cite{DBLP:conf/icassp/LohiaRBSVP19,DBLP:conf/sdm/FishKL16,DBLP:conf/icdm/KamiranCP10}). It is not usually specified which scenarios each method may be applied in ~\cite{DBLP:journals/jair/Weinberg22,DBLP:journals/clsr/WachterMR21} and there is usually no consideration into the legality of the use of protected attributes for the identification of bias~\cite{DBLP:conf/ssci/HaeriZ20}.

Most existing bias detection and mitigation methods focus on group fairness and do not consider individual fairness. Satisfying individual fairness in a decision-making system is necessary but not sufficient to ensure overall fairness~\cite{DBLP:conf/aies/Fleisher21}. This is because individuals in a protected group could all be given negative outcomes and this would conflict with the notion of group fairness despite individual fairness being satisfied~\cite{DBLP:conf/kbse/ChakrabortyPM20}.

\subsection{Computational Argumentation \label{argumentation}}

Argumentation theory has been explored as a way of representing arguments and forms of reasoning \cite{DBLP:journals/ai/Dung95,DBLP:journals/ker/BaroniCG11,Atkinson:17}. An Abstract Argumentation Framework (AAF) is a tuple $\langle A, R\rangle$, where $A$ is a set of arguments and $R \subseteq A\times A$ is an attack relation between them. Arguments are atomic entities whose content is not specified, the focus being on acceptability criteria for the arguments based on the attack relationship, resulting in alternative semantics \cite{DBLP:phd/ethos/Efstathiou11,DBLP:journals/ker/BaroniCG11}.

Weighted (or Quantitative) Argumentation Frameworks (WAFs) augment AAFs with {\em weights}. For our method, we are interested in WAFs that define weights for both arguments and attacks.

\begin{definition}[Weighted Argumentation Graph \cite{DBLP:conf/atal/AmgoudD19}] $G = \langle A, \sigma, R, \pi\rangle$, where $A$ is a non-empty finite set of
arguments, $R \subseteq A \times A$, $\sigma : A \mapsto [0,1]$, and $\pi : R \mapsto [0, 1]$.
\end{definition}

$A$ and $R$ define arguments and attacks as before, but an argument $a \in A$ is given an initial weight $\sigma(a)$, and attacks $(a,b)$ between arguments are given a strength $\pi(a,b)$. Semantics for WAFs define how the final weights of the arguments are calculated given $G$. We use the strength of attacks to express the relative representation of ``votes'' by similar individuals, inspired by ideas from \citet{leite-martins:13,gabbay-rodrigues-jlc:13}.
WAFs may have cycles, and applicable semantics include the Trust-based semantics~\cite{DBLP:conf/ijcai/PereiraTV11}, the Simple Product~\cite{DBLP:conf/ijcai/LeiteM11}, Weighted Max-based \emph{(Mbs)}, Weighted Cardinality-based \emph{(Wbs)} and Weighted h-Categorizer (\emph{Hbs})~\cite{DBLP:journals/ai/AmgoudDV22}. These last three were originally developed for frameworks with attack and support relations~\cite{DBLP:conf/ijcai/AmgoudBDV17} and have also been extended to account for strengths of attacks~\cite{DBLP:conf/atal/AmgoudD19}.
\emph{Mbs} favours the strength over quantity of attacks, whereas \emph{Cbs} favours quantity over strength. \emph{Hbs} considers both~\cite{DBLP:conf/atal/AmgoudD19}.

\emph{Hbs} is based on the h-Categorizer semantics originally developed for non-weighted graphs~\cite{DBLP:journals/ai/BesnardH01,DBLP:conf/ksem/PuLZL14}. \emph{Hbs} defines an infinite sequence of weights for arguments $s(a)^{(0)}, s(a)^{(1)}, \ldots$ such that
\begin{eqnarray}
\!\!\!\!\!\!\!s(a)^{(1)}\!\!\! & = & \!\!\!\sigma(a)\\
\!\!\!\!\!\!\!s(a)^{(n+1)}\!\!\! & = & \!\!\!\frac{\sigma(a)}{1+\sum\limits_{b \in Att_{a}} \pi((b,a)) \cdot s(b)^{(n)} }\text{, for $n > 0$}\label{eq:weight-updates}
\end{eqnarray}

Notice that the denominator of the fraction on the right-hand side of Equation~\ref{eq:weight-updates} is greater than $1$ for any {\em attacked} argument $a$ provided the strength of the attack is non-null. This means the weight of such arguments decrease in proportion to the quantity as well as the strength of the attacks. The sequence $\{s(a)^{(n)}\}^{+\infty}_{n=1}$ is infinite. However, \citet[Theorem~17]{DBLP:journals/ai/AmgoudDV22} showed that for every $a \in A$, it  converges and the final weight of an argument $a$ is defined as $\lim_{n\to + \infty} s(a)^{(n)}$. We will describe how to approximate these final weights in Section~\ref{method}.

\section{Identifying Reasons for Bias using Argumentation\label{method}}

In this section we describe our method. Given a \emph{queried} individual's classification, we aim to ensure it is fair with respect to similar individuals. For simplification, we only consider queried individuals that are classified \emph{negatively}, although our method equally applies to positive classifications.
Our approach requires a set of unlabelled individuals, which we obtain from the test data (see Section~\ref{sec:experiments}).

Let $E$ be the set of unlabelled individuals with attributes $Z=\langle z_1, z_2,...,z_p\rangle $ and corresponding domains $\langle D_1, D_2, ..., D_p\rangle $. Let $v: E \times Z \rightarrow \cup_{i=1}^p D_i$ be the function that given an individual $e$ and
attribute $z_i$, returns the value of the attribute $z_i$ for $e$.

\subsection{Defining Similar Individuals \label{similar}}
\newcommand{\sind}[2]{\text{$Sim_{#1}(#2)$}}
\newcommand{\rbit}[1]{{\color{red}{\bf REMOVE:}#1}}

To represent individual fairness, we must first define what it means for individuals to be ``similar''. Several definitions of similarity have been explored in the literature in relation to individual fairness. For example, \citet{DBLP:conf/icml/ZemelWSPD13} proposed the use of k-nearest neighbours (KNN)  while \citet{DBLP:conf/innovations/DworkHPRZ12} proposed context-specific similarity definitions whereby certain attributes are considered more important in the evaluation of how similar individuals are to one another.

We identify an individual $e$'s similar individuals as the nearest neighbours in $E/\{e\}$ using a Ball Tree clustering algorithm \cite{DBLP:conf/bmvc/LeibeMS06} using the Hamming distance. We assume all values to be categorical, converting numerical attributes such as \emph{age} to categorical attributes by grouping them appropriately (see Section~\ref{sec:experiments}).\footnote{The distance between individuals could also be calculated using another metric that would consider the difference between numerical values and/or the relative distance between ordered categorical attributes such as $\dash{education-level}$.}

Hence, the distance $d(e_1,e_2)$ between individuals $e_1$ and $e_2$ is given by
$$d(e_1,e_2)=\sum_{i = 1}^{p} d_i, \text{~where~} d_i\! =\!
\begin{cases}
  0,\!\!\!\!\! & \text{if~} v(e_1, z_i) = v(e_2, z_i) \\
  1,\!\!\!\!\! & \text{if~} v(e_1, z_i) \neq v(e_2, z_i)\\
\end{cases} $$
The distance between individuals with the same attribute values is $0$ and that the distance between two individuals increases in proportion to the number of attributes with different values in the individuals.


\sind{k}{e} will denote the set of $k$ individuals most similar to the individual $e$, i.e.
the individuals $\{e_1,\ldots,e_k\} \subseteq E / \{e\}$ minimising $\sum_{i=1}^{k} d(e,e_i)$. We assume that $|E/\{e\}|\geq k$ and that if there are more than $k$ individuals minimising $\sum_{i=1}^{k} d(e,e_i)$, we can choose arbitrarily between them.

In Section \ref{sec:experiments} we discuss the effect of using different values of $k$. In what follows, we present a working example to illustrate our method using $k=5$ similar individuals.

\begin{example} \label{example:bias}

Table~\ref{tab:exampledata1} shows six individuals from the commonly used Adult dataset~\cite{misc_adult_2}, restricted to three attributes and their classifications. The top row represents the queried individual $e$ and the other rows represent $e$'s similar individuals as described in this section. The classification $+$ represents that the individual's income is predicted to exceed \$50K/year, and $-$ otherwise.

\begin{table}[ht]
\centering
{\small
\begin{tabular}{l|l|l|c}
\textbf{workclass} & \textbf{education} & \textbf{race} & \textbf{Classification} \\ \hline \hline
\rowcolor{light-gray}
 $\dash{Local-gov}$  & $Bachelors$          & $Black$         & $-$      \\ \hline
$Private$           & $Bachelors$             & $White$         & $+$    \\ \hline
 $\dash{Local-gov}$          & $\dash{HS-grad}$          & $White$         & $+$    \\ \hline
 $\dash{Local-gov}$         & $Bachelors$           & $White$         & $+$   \\ \hline
 $Private$             & $Masters$            & $White$         & $+$  \\ \hline
 $\dash{Local-gov}$         & $Masters$        & $White$         & $+$    \\
\end{tabular}
}
\caption{Sample data with queried individual (grey) and five similar individuals.}
\label{tab:exampledata1}
\end{table}

\end{example}

Intuitively, Example~\ref{example:bias} shows that given the queried individual in the top row and its similar individuals, $(race, Black)$ is the attribute-value pair contributing the most to the negative classification of the queried individual, since all combinations of values of the other attributes in the similar individuals lead to a positive classification. Our aim is to create a mechanism by which this type of bias can be identified, focusing on \emph{why} an individual has been classified negatively (and which attributes contribute most to this), as opposed to \emph{how many} individuals have been treated unfairly which is the aim of most existing bias detection methods~\cite{waller2023bias}.

\subsection{Constructing the Argumentation Graph \label{construct}}

Let $f$ be a binary classifier $f:E \rightarrow Y$ which takes an individual $e \in E$ as input and outputs a classification $f(y) \in \{+,-\}$. Recall that $v(e, z)$ is the value of the attribute $z$ for $e$. We use $e_0$ to denote the queried individual.

We show how to construct the weighted argumentation graph $\langle A,\sigma,R,\pi \rangle$, which will be used to detect the attribute-value pairs that contribute the most to the classification of $e_0$. In all that follows, we assume that the set of attributes is $Z=\{z_1,\ldots,z_p\}$.

We define the set of arguments $A$ as the unique attribute-value pairs $(attribute, value)$ representing an association $attribute = value$ found in $\{e_0\} \cup \sind{k}{e_0}$.

\begin{definition}[Set of arguments] \label{def:args}
    Let $e_0$ be the queried individual and $\sind{k}{e_0}$ the set with the $k$ individuals most similar to $e_0$ according to some similarity measure $Sim$. The set of arguments $A$ is defined as follows. $$A = \bigcup_{i=0}^k \bigcup_{j=1}^{p} \{(z_j,v(e_k,z_j))\}$$
\end{definition}

At the outset, we have no prior information about which attribute-value pairs contribute the most to the negative classification of the queried individual, so the initial weight of all arguments is set to $1$.\footnote{It is left for future work to explore whether the initial weight of an argument could represent prior knowledge such as prevalence of attribute values in the dataset or existing group fairness metrics for a deployed model.}

\begin{definition}[Initial argument weights] For all $a \in A$, $\sigma(a)=1$.
\label{def:init-weights}
\end{definition}

We now define the attack relations and strengths using the frequency of occurrence of those pairs amongst the similar individuals. One way of looking at this is that individuals ``argue'' for the potential reasons of the negative classification of the queried individual, which can be the presence of one, or multiple attribute-value pairs. If a similar positively-classified individual has a value for an attribute $z$ different to the value of the negatively-classified queried individual, then the similar individual attacks that attribute-value pair in the queried individual. This will in turn reduce the initial weight of the pair. Specifically {\em all} attribute-value pairs of  similar individuals $e_i$ will attack the value of the attribute $z$ in the queried individual $e_0$ if the values of $z$ differ in $e_0$ and $e_i$. This accounts for the possibility that $e_0$'s negative classification could be due to combinations of multiple attribute-value pairs in the similar individuals.

\begin{definition}[Attack relationship] \label{def:addattacks}
    Given a binary classifier $f:E \rightarrow \{+,-\}$, a queried individual $e_0$, and the set of similar individuals $\sind{k}{e_0}$:
    \begin{multline*}
        R = \bigcup_{i=1}^k \bigcup_{j=1}^p  \bigcup_{l=1}^p \{((z_l,v(e_i,z_l)), (z_j,v(e_0,z_j))) \;|\\ f(e_i) \neq f(e_0) \text{~and~} v(e_0,z_j) \neq v(e_i,z_j)\}
    \end{multline*}
\end{definition}

Notice that by Definition~\ref{def:addattacks} the targets of attacks can only be arguments of the type $(z_j,v(e_0,z_j))$ (i.e. towards the queried individual's attribute-value pairs) and that attacks from the attribute-value pairs of a similar individual $e_i$ only occur towards an argument $(z_j,v(z_j,e_0))$ when the values of the attribute $z_j$ differ in $e_0$ and $e_i$ (i.e. ($v(e_0,z_j) \neq v(e_i,z_j)$). In addition, we also want the strength of any potential attacks to reflect the {\em proportion} of similar individuals with a particular differing attribute-value pair. This approach is inspired by voting, with  related ideas explored in detail in \citet{leite-martins:13,gabbay-rodrigues-jlc:13}. The following definition captures the intuition.

\begin{definition}[Strength of attacks] Take $(a,b) \in R$, where $a=(z_1 ,v_1)$ and $b=(z_2,v_2)$.
\[\pi((a,b))=\frac{|\{e_i\, |\, v(e_i,z_1 )=v_1 \text{~and~} v(e_0,z_2)\neq v(e_i,z_2)\}|}{k}\]
\label{def:att-strength}
\end{definition}

Since $(a,b) \in R$, there are between $1$ and $k$ individuals for which the condition in Definition~\ref{def:att-strength} holds. Therefore $\frac{1}{k}\leq \pi((a,b)) \leq 1$.

\begin{example} \label{example:attacks}
Take the attribute {\em workclass} in Table~\ref{tab:exampledata1} for which two similar individuals have the value $Private$, different to the queried individual. There is an attack relation from all attributes in these two similar individuals to $(workclass, \dash{Local-gov})$. The values of attribute $education$ for these individuals are $Bachelors$ and $Masters$ therefore an attack is added from $(education,Bachelors)$ and $(education,Masters)$ to $(workclass, \dash{Local-gov})$, both with strength $\frac{1}{5}$ according to Definition~\ref{def:att-strength}. $(race,White)$ also attacks $(workclass, \dash{Local-gov})$ but both individuals have that attribute value so the strength of the attack is $\frac{2}{5}$. Similarly for $(workclass, Private)$ to $(workclass, \dash{Local-gov})$.

\begin{figure}[ht]
    \centering
    \hspace*{5pt}\resizebox{.9\linewidth}{!}{
        \begin{tikzpicture}[
    every matrix/.style={ampersand replacement=\&,column sep=13mm,%
      row sep=1cm,inner sep=0pt},
    every loop/.style={min distance=6mm,in=330,out=210,looseness=8},
    every node/.style={align=center},
    every edge quotes/.append style={font=\normalsize, align=center, auto}
    ]

  \matrix{
    \node[barg,text width=18mm, label={$s^{\varepsilon}=0.48$}] (lg) {\small $(workclass,$\\$\dash{Local-gov})$};\&
    \node[barg,text width=18mm, label={$s^{\varepsilon}=0.39$}] (ba) {\small $(education,$ \ $Bachelors)$};\&
    \node[barg,text width=18mm, label={\bm{$s^{\varepsilon}=0.29$}}] (bl) {\small\begin{tabular}{c} $(race,$\\$Black)$ \end{tabular}};\\
    \node[barg,text width=18mm, label=below:{$s^{\varepsilon}=1$}] (pr) {\small$(workclass,$ \ $Private)$};\&
    \node[barg,text width=18mm, label=below:{$s^{\varepsilon}=1$}] (ma) {\small$(education,$ \ $Masters)$};\&
    \node[barg,text width=18mm, label=below:{$s^{\varepsilon}=1$}] (wh) {\small $(race,$\\$White)$};\\
  };

    \node [barg,text width=18mm, label=below:{$s^{\varepsilon}=1$}, below of=ma, yshift=-1.3cm] (hs) {\small $(education,$ \ $\dash{HS-grad})$};

  \draw[very thin, -{Latex[width=2mm,length=2mm]}]
  (pr) edge["\normalsize $\frac{2}{5}$", pos=0.2, outer sep=-2pt] (lg)
  (pr) edge["\normalsize $\frac{1}{5}$", pos=0.1, outer sep=-2pt] (ba)
  (pr) edge["\normalsize $\frac{2}{5}$", pos=0.05, outer sep=-2pt, swap] (bl)
  (ma) edge["\normalsize $\frac{1}{5}$", pos=0.1, outer sep=-2pt] (lg)
  (ma) edge["\normalsize $\frac{2}{5}$", pos=0.2, outer sep=-2pt] (ba)
  (ma) edge["\normalsize $\frac{2}{5}$", pos=0.1, outer sep=-2pt, swap] (bl)
  (wh) edge["\normalsize $\frac{2}{5}$", pos=0.03, outer sep=-2pt] (lg)
  (wh) edge["\normalsize $\frac{3}{5}$", pos=0.1, outer sep=-2pt, swap] (ba)
  (wh) edge["\normalsize $\frac{5}{5}$", pos=0.2, outer sep=-2pt, swap] (bl)
  (hs) edge["\normalsize $\frac{1}{5}$", pos=0.05, outer sep=-2pt, bend right=45] (ba)
  (hs) edge["\normalsize $\frac{1}{5}$", pos=0.02, outer sep=-2pt, swap, bend right=80] (bl)
  (lg) edge["\normalsize $\frac{2}{5}$", pos=0.15, outer sep=-2pt, bend left=5] (ba)
  (lg) edge["\normalsize $\frac{3}{5}$", pos=0.1, outer sep=-2pt, bend left=40] (bl)
  (ba) edge["\normalsize $\frac{1}{5}$", pos=0.15, outer sep=-2pt, bend left=5] (lg)
  (ba) edge["\normalsize $\frac{2}{5}$", pos=0.1, outer sep=-2pt] (bl);

\end{tikzpicture}}
    \caption{The weighted argumentation graph constructed from Table~\ref{tab:exampledata1} with the final weights $s^\varepsilon$ of all arguments.\label{fig:example_final}}
\end{figure}
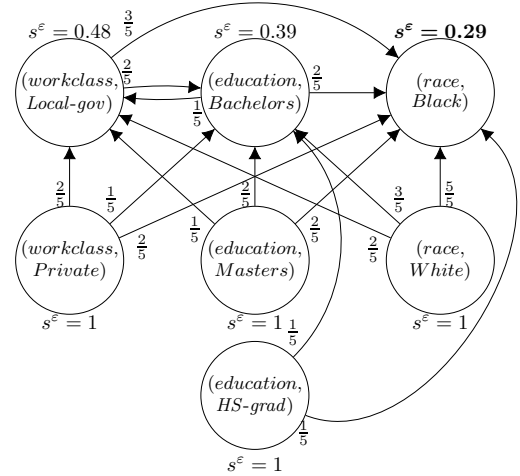

\end{example}

Figure~\ref{fig:example_final} shows the weighted argumentation graph constructed from the data in Table~\ref{tab:exampledata1}, with the strengths of attacks between arguments, and the final weights of all arguments, which will be discussed next.

\subsection{Calculating Final Weights of Arguments}

We now need to calculate the arguments' final weights, using an appropriate well-defined semantics. Good candidates for semantics are those that aggregate attacks in an ``argumentation-friendly'' way (see \citet{esnan:15} for a discussion). At the very least, the semantics needs to be able to deal with potential cycles in the graph and reduce initial weights in proportion to the strength of attacks on arguments. We chose the \emph{Hbs} semantics not only because it satisfies these requirements, but it is also efficient to compute and provided good results with the datasets used in our experiments. Further properties of this semantics have been explored in~\citet{DBLP:conf/atal/AmgoudD19}.\footnote{We also ran experiments using the Quadratic Energy Model semantics~\cite{DBLP:conf/kr/Potyka18} (without supports) and it identified the same weakest arguments for all individuals.
}

\newcommand{\final}[1]{\text{$s(#1)^{\varepsilon}$}}

Recall Equation~\ref{eq:weight-updates}. We aim to approximate $\lim_{n\to + \infty} s(a)^{(n)}$ by successively calculating the values of the sequence until the difference between successive values of all arguments is within a desired convergence threshold $\varepsilon$, i.e. $s(a)^{(n+1)} - s(a)^{(n)} < \varepsilon$, for all $a \in A$.
By an abuse of notation we will denote this value $\final{a}$ and call it the ``final'' weight of the argument $a$ (up to $\varepsilon$). Section~\ref{sec:experiments} describes the particular choice of $\varepsilon$ used in the experiments.

Having computed all final weights, the weakest arguments, i.e. those with the smallest final weights, correspond to the attribute-value pairs of the queried individual that most contribute to its negative classification. These are given as the explanation $Exp$ for the classification: $Exp = \{a \in A\;|\;\final{a}=\min_{b \in A}\{\final{b}\}\}$.

The approximated final weights calculated using \emph{Hbs} for the sample data in Table~\ref{tab:exampledata1} are shown in Figure~\ref{fig:example_final}. It is easy to see that $(race,Black)$ is the weakest argument in the graph, supporting our intuition in Example~\ref{example:bias}. We now show general properties of the final weights.

\begin{proposition}
Let $a=(z_i, v)\in A$.
If $\final{(z_i, v)} < 1$, then $v(e_0,z_i) = v$, where $e_0$ is the queried individual.
\begin{proof}
According to Definition~\ref{def:init-weights}, $s((z_i, v))^{(1)}=\allowbreak\sigma((z_i, v))=\allowbreak1$. According to Equation~\ref{eq:weight-updates}, argument weights can only decrease and only when an argument is attacked. According to Definition~\ref{def:addattacks}, attacks are only defined towards attribute-value pairs of the queried individual. If $\final{(z_i, v)} < 1$, then $(z_i, v)$ must have been attacked, and hence $(z_i, v)$ is in the queried individual.
\end{proof}
\end{proposition}

\begin{proposition} \label{prop:consistent}
Assume the queried individual $e_0$ has the same classification as all of its similar individuals. Let $a_i$ be the attribute-value pair ($z_i,v_i$), such that $v_i=v(e_0,z_i)$, then $\final{a_i}=\sigma(a_i)=1$.

\begin{proof}
Let the argument $a_j$ be an attribute-value pair $(z_j,v_j)$, such that $v_j = v(e_0,z_j)$. According to Definition~\ref{def:addattacks}, an attack is added into $a_j$, only when the classification of $e_0$ is different to the classification of a similar individual $e_i$. Since the classification of $e_0$ is the same as that of all of its similar individuals, there will be no attacks into any such $a_j$, and hence according to Equation~\ref{eq:weight-updates}, $s(a_j)^{n+1}=s(a_j)^{1}=\final{
a_j}$, for all $1 \leq j \leq p$, 
$n\geq 0$.
\end{proof}
\end{proposition}

Proposition~\ref{prop:consistent} shows that when the classification of the queried individual $e_0$ matches  the classifications of its similar individuals, no attribute-value pair is singled-out in $e_0$.

\begin{proposition}
Assume there is only one attribute $z$ for which all positively-classified similar individuals have a different value than the queried individual and all other attributes have the same values for all individuals. Let $a$ be the attribute-value pair $(z, v(e_0,z))$, then $\final{a} < 1, \final{a} = \min_{b \in A}\{\final{b})\}\}$ and for all $b \in \{A \setminus a\}, \final{b} = 1$.

\begin{proof}
According to Definition~\ref{def:addattacks}, $(z, v(e_0,z))$ is attacked by all attribute values in all positively-classified similar individuals. Since all other attribute-value pairs are the same, then $(z,v(e_0,z))$ is the only argument that is attacked.

According to Equation~\ref{eq:weight-updates}, $s(a)^{n+1}$ only decreases if $\{ b \, |\, (b,a) \in R\} \neq \emptyset$ thus only $(z,v(e_0,z))$ can have final weight less than $1$ (all other final weights remaining unchanged, i.e. equal to $1$). Hence, $\final{a}$ is the minimum value and identified as the only possible explanation for the negative classification.
\end{proof}
\end{proposition}

\section{Evaluation\label{sec:evaluation}}
\label{sec:experiments}

We conduct our analysis with experiments on the Adult~\cite{misc_adult_2} and the Bank Marketing ~\cite{misc_bank_marketing_222} datasets, commonly used in the fairness literature. Further, we introduce bias to the Adult dataset to illustrate our method identifies bias as expected.

\subsection{Experiments on Real Data}

We cannot easily compare our results to existing fairness metric values or bias detection methods as we focus on the reasons for a classification, as described in Section~\ref{method}, which provides the basis for {\em explaining} a classification. We envision our method being employed to identify possible bias in an individual's classification (see Example~\ref{example:bias}), rather than for a group of individuals. Nonetheless, we offer some results to showcase the efficacy of our approach.

To be able to show meaningful results, we first provide a qualitative data analysis of the Adult and Bank Marketing datasets, with a focus on their protected attributes, even though our method does not require their specification. We then provide the results of our experiments.

\subsubsection{Qualitative data analysis} \label{sec:data-analysis}

The original Adult dataset has 48,842 instances. We pre-process to remove instances with null values to result in 45,222 instances. For this dataset, the attributes $sex$ and $race$ are identified as protected throughout the literature~\cite{DBLP:journals/widm/QuyRIZN22}. Table~\ref{tab:adult-prev} shows the prevalence of positive and negative labels for the values of these attributes. A negative (positive) label represents the fact that an individual's income is below (above) \$50K, respectively. The percentage of negative labels is the proportion of individuals in a group with a negative label out of all individuals in that group. Table~\ref{tab:adult-prev} shows there is a greater percentage of negative labels for females than males and for non-white individuals (except $AsiaPacIslander$) compared to white individuals.

\begin{table}
\centering
{\small
\begin{tabular}{@{\hspace*{3pt}}c|@{\hspace*{3pt}}l@{\hspace*{3pt}}|@{\hspace*{3pt}}r@{\hspace*{3pt}}|@{\hspace*{3pt}}r@{\hspace*{3pt}}|@{\hspace*{3pt}}r@{\hspace*{3pt}}}

& \textbf{Attribute value} & \textbf{$+$ labels} & \textbf{$-$ labels} & \textbf{\% of $-$ labels} \\\hline
\multirow{2}{*}{\rotatebox[origin=c]{90}{sex}} & Male &  9539 & 20,988 & 69\% \\
& Female & 1669 & 13,026 & 89\% \\ \hline

\multirow{5}{*}{\rotatebox[origin=c]{90}{race}} & White & 10,207 & 28,696 & 74\% \\
& Black & 534 & 3694 & 87\% \\
& AsianPacIslander & 369 & 934 & 72\% \\
& AmerIndianEskimo  & 53 & 382 & 88\% \\
& Other & 45 & 308 & 87\%
\end{tabular}
}
\caption{Prevalence of positive and negative labels for different protected attribute values in the Adult dataset.}
\label{tab:adult-prev}\end{table}

The Bank Marketing dataset has 45,211 instances and the attributes $age$ and $marital$ are identified as protected. We pre-process the dataset to categorise the attribute $age$ into two groups corresponding to the protected groups identified in the literature~\cite{DBLP:journals/widm/QuyRIZN22}. Specifically, converting the attribute value of $age$ to $YoungOrOld$ where it is less than 25 or greater than 60 and $MidAge$ otherwise.

It is not as clear what a positive label is for this dataset. We define a positive (negative) label as representing individuals that have not (have) subscribed to a term deposit. As previously mentioned, we could easily adapt our method to identify bias with respect to a positive classification. Table~\ref{tab:bank-prev} shows the prevalence of positive and negative labels for the values of these attributes.

For the attribute $age$, individuals with value $YoungOrOld$ are considered in the protected group, whereas the individuals with value $MidAge$ are not. For the attribute $marital$, non-married individuals with the values of $single$ or $divorced$ are considered in the protected group, whereas $married$ individuals are not. Table~\ref{tab:bank-prev} shows that there is a greater percentage of negative labels for young or old individuals than mid-age individuals, and for single and divorced individuals compared to married individuals, however the percentage of negative labels does not vary greatly for $marital$.

\begin{table}
\centering
{\small
\begin{tabular}{c|l|r|r|r}

& \textbf{Attribute value} & \textbf{$+$ labels} & \textbf{$-$ labels} & \textbf{\% of $-$ labels} \\\hline

\multirow{2}{*}{\rotatebox[origin=c]{90}{age}} & MidAge & 38,634 & 4580 & 11\% \\
& YoungOrOld & 1288 & 709 & 36\% \\ \hline

\multirow{3}{*}{\rotatebox[origin=c]{90}{marital}} & Married & 24,459 & 2755 & 10\% \\
& Single & 10,878 & 1912 & 15\% \\
& Divorced & 4585 & 622 & 12\% \\
\end{tabular}
}
\caption{Prevalence of positive and negative labels for different protected attribute values in the Bank Marketing dataset.}
\label{tab:bank-prev}
\end{table}

\subsubsection{Setup}

For our experiments, we pre-process the Adult dataset as before by removing null values. Additionally we remove the attributes $fnlwgt$\footnote{$fnlwgt$ represents a weighting of how many individuals an instance represents and is not a feature of an individual (commonly removed from the dataset~\cite{DBLP:journals/kais/KamiranC11}).} and $\dash{education-num}$\footnote{$\dash{education-num}$ is equivalent to $\dash{education-level}$ which is already included in the dataset.}. We then apply the same numerical attribute categorisation as outlined by \citet{DBLP:journals/widm/QuyRIZN22} to both datasets. This aligns with our similarity definition in Section~\ref{similar} and ensures meaningful argument representations.

To evaluate our method, we split our pre-processed datasets into training (80\%) and test (20\%) partitions and train a logistic regression classifier on the training sets. We chose logistic regression because it is the most common classification model used to evaluate bias mitigation methods \cite{DBLP:journals/corr/abs-2207-07068}. However, our method is model-agnostic and can be applied with any classifier.

We took as queried individuals every individual with a negative classification from our test sets amounting to 7,252 and 1,253 individuals for the Adult and Bank Marketing datasets, respectively. We then selected the $5$ most similar individuals in the test data using a Ball-tree clustering algorithm as defined in Section~\ref{similar}.\footnote{In our experiments, the KD clustering algorithm had similar performance to Ball-tree clustering.} This allowed us to collate groups of individuals such as those in Table~\ref{tab:exampledata1} and use Definitions~\ref{def:args}~--~\ref{def:addattacks} to construct the weighted argumentation graph such as in Figure~\ref{fig:example_final}. We then calculated the strengths of the attacks according to Definition~\ref{def:att-strength} and used the \emph{Hbs} semantics (Equation~\ref{eq:weight-updates}) to find the weakest arguments in the graphs.

\begin{table}
\centering
{\small
\begin{tabular}{l|c|c|c|c}
 \textbf{Dataset} &  \textbf{Train} & \textbf{Test} & \textbf{Accuracy} & $\mathbf{F_{1}}$ \\ \hline
Adult & 36177 & 9045 & 85\% & 66\% \\ 
Bank Marketing & 36169 & 9042 & 72\% & 82\%
\end{tabular}
}
\caption{Number of instances in the train and test set for the Adult and Bank Marketing datasets, with classifier performance statistics.}
\label{tab:data-count}
\end{table}

We tested varying convergence thresholds ($\varepsilon$) for approximating final weights, as in Equation~\ref{eq:weight-updates}. We chose $\varepsilon = 0.01$; smaller values did not impact weakest argument identification but increased computation time. Larger $\varepsilon$ values led to imprecise weight approximations that could not reliably identify the weakest arguments. All approximated weights are rounded to two decimal places, treating two argument weights as equal if their rounded values match.

As shown by Proposition~\ref{prop:consistent}, if all similar individuals have the same (negative) classification as a queried individual, the final weights of all arguments are 1 and hence the queried individual has been treated consistently with respect to the similar individuals. Otherwise the weakest arguments will correspond to the attribute-value pairs contributing the most to the negative classification.

\subsubsection{Results on the Adult dataset}
Table~\ref{tab:adult-results} shows the count and proportion of where a protected attribute value was amongst the weakest arguments in the graph. Furthermore, 70\% of the queried individuals are consistent with the similar individuals, meaning all similar individuals are also classified negatively. Our method identified 8.2\% of queried individuals of which $(sex,Female)$ contributed the most to the negative classification, thus identifying bias against these individuals. As expected from the prevalence of negative labels for females versus males, this is greater than the proportion of $(sex,Male)$ contributing the most to the negative classification. This detects individuals who have been given a negative classification unfairly based on the attribute values $Male$ or $Female$, highlighting a benefit of our method in not requiring to specify the protected group before deployment.

Similarly, Table~\ref{tab:adult-results} shows that $(race, Black)$ was identified as the attribute-value contributing the most to the negative classification in 159 more cases than the attribute-value $(race, White)$. Summing the proportions for all values of $race$ that are not equal to $White$, we obtain 5.0\%, showing that there are a greater percentage of non-white individuals being negatively-classified due to their value of $race$ than white individuals.

\begin{table}
\centering
{\small
\begin{tabular}{c|l|r}
& \textbf{Attribute value} & \textbf{Count (proportion)}\\
\hline
\multirow{2}{*}{\rotatebox[origin=c]{90}{sex}} & Male & 130 (1.8\%) \\
& Female & 595 (8.2\%) \\
\hline
\multirow{5}{*}{\rotatebox[origin=c]{90}{race}} & White & 86 (1.2\%) \\
& Black & 245 (3.4\%) \\
& AsianPacIslander & 64 (0.9\%) \\
& AmerIndianEskimo & 28 (0.4\%) \\
& Other & 21 (0.3\%) \\
\end{tabular}
}
\caption{Count (and proportion) of attribute values being the weakest argument in queried individuals (Adult).}
\label{tab:adult-results}
\end{table}

\subsubsection{Results on the Bank Marketing dataset}  Table \ref{tab:bank-results} shows the count and proportion of the weakest arguments for all the protected attribute values. Furthermore, 21\% of the queried individuals are consistent with similar individuals, hence all similar individuals are also classified negatively.

Our method identifies 8.9\% of queried individuals of which $(age, YoungOrOld)$ contributes the most to the negative classification, thus identifying bias against these individuals. As expected from the prevalence of negative labels for $MidAge$ versus $YoungOrOld$, this is less than the proportion of $(age,MidAge)$ contributing the most to the negative classification.

The counts of weakest arguments for various $marital$ values exhibit no significant difference, much like the proportion of negative labels for each group in the original dataset.

\begin{table}
\centering
{\small
\begin{tabular}{c|l|r}
& \textbf{Attribute value} & \textbf{Count (proportion)}\\
\hline
\multirow{2}{*}{\rotatebox[origin=c]{90}{age}} & MidAge  & 14 (1.1\%) \\
& YoungOrOld & 112 (8.9\%) \\
\hline
\multirow{3}{*}{\rotatebox[origin=c]{90}{marital}} & Married & 59 (4.7\%) \\
& Single & 68 (5.4\%) \\
& Divorced & 51 (4.0\%) \\
\end{tabular}
}
\caption{Count (and proportion) of attribute values being the weakest argument in queried individuals (Bank Marketing).}
\label{tab:bank-results}
\end{table}

\subsection{Experiments on Adapted Data} \label{sec:adapteddata}

To provide evidence our method detects bias as expected, we add an attribute $\dash{bias-attr}$ to the test set of the Adult dataset, which we first pre-process as previously described. The attribute $\dash{bias-attr}$ takes value 0 or 1, each with probability $0.5$. We then fix the classifier $f:E \rightarrow Y$ such that $f(e)=v(e,\dash{bias-attr})$.

We hypothesise that all queried individuals will either be consistent with their similar individuals, i.e. there is no difference in classifications between a queried individual and its similar individuals, or $\dash{bias-attr} = 0$ will be identified as contributing the most to the negative classification of the queried individual.

\subsubsection{Results}

Our method correctly identifies that, for all negatively-classified queried individuals where at least one of the similar individuals is positively-classified, $\dash{bias-attr}=0$ is amongst the weakest arguments and therefore identified as contributing the most to the negative classification, i.e. $\dash{bias-attr}=0$ is amongst the weakest arguments in our constructed argumentation graph. This is the case for 56.4\% of the queried individuals.

The other 43.6\% of individuals were consistent with similar individuals thus no bias was detected. This highlights a limitation of individual fairness --- if all similar individuals have the same classification, there is no bias identified as they are treated the same, even if that they are all classified negatively due to a particular attribute value. By increasing the number of similar individuals we consider, we tend towards group fairness and the proportion of consistent queried individuals decreases. For example, running the experiments with \emph{10 similar individuals}, we identify 76.8\% of queried individuals for which the reason for the classification is $\dash{bias-attr}=0$ and 85.3\% with \emph{15 similar individuals}.

Our method correctly identifies the attribute value that contributes to the negative classification for all individuals that have at least one similar individual with a different classification. Ideally, dataset curation should prioritise minimising strong correlations between protected attributes and labels. This can be achieved through the implementation of bias mitigation techniques that emphasise group fairness. Assuming this precondition, our method excels in uncovering less obvious biases associated with individual fairness, which is often overlooked in practice.

\section{Conclusion \& Future Work \label{conclusion}}

In this paper, we proposed a novel argumentation-based method for finding why an individual is classified differently from similar individuals, to be able to identify bias in relation to individual fairness. Our method is model-agnostic and does not require access to labelled data or the specification of protected characteristics.

We construct a quantitative argumentation framework based on the relationships between attribute-value pairs of a queried individual and its similar individuals. The argumentation framework can be used to extract an \emph{explanation} for the classification of the queried individual compared to its neighbours, thus offering a transparent representation of the reasons for a classification, which is applicable to any machine learning classifier.

We evaluated our method in the Adult and Bank Marketing datasets, commonly used in the fairness literature. Our results demonstrate a correlation between the attribute-value pairs detected by our method and the prevalence of protected attributes in the original dataset. In addition, we introduced synthetic bias into the Adult dataset and showed that our method correctly identified the attribute-value pairs artificially responsible for the classification.

Identifying the reasons for an individual's particular black-box binary classification is the first step to creating more transparent methods whose results can be better explained to users, hence increasing the trustworthiness of the classification.

There are multiple avenues for future work. Although we identified the reasons for bias in the datasets analysed, we would like to extend our experiments to cover other datasets and ensure that the current semantics is sufficiently fine-grained to identify potential more subtle causes for bias. In addition, we plan to employ alternative definitions of individual similarity and consider and compare the results obtained using different argumentation semantics.

Future work also includes the development of templates that can be instantiated with more intelligible explanations in natural language rather than the current identification of attribute-value pairs, thus allowing better understanding of the explanations by end-users.

\appendix

\section{Acknowledgments}

This work was supported by the UK Research and Innovation Centre for Doctoral Training in Safe and Trusted Artificial Intelligence [grant number EP/S023356/1]\footnote{\url{www.safeandtrustedai.org}}. The first author is an Affiliate of the King's Institute for Artificial Intelligence and additionally funded by The Alan Turing Institute’s Enrichment Scheme.

\bibliography{ref}

\begin{thebibliography}{57}
\providecommand{\natexlab}[1]{#1}

\bibitem[{Amgoud et~al.(2017)Amgoud, Ben{-}Naim, Doder, and
  Vesic}]{DBLP:conf/ijcai/AmgoudBDV17}
Amgoud, L.; Ben{-}Naim, J.; Doder, D.; and Vesic, S. 2017.
\newblock Acceptability Semantics for Weighted Argumentation Frameworks.
\newblock In \emph{Proceedings of the Twenty-Sixth International Joint
  Conference on Artificial Intelligence, {IJCAI} 2017, Melbourne, Australia},
  56--62.

\bibitem[{Amgoud and Doder(2019)}]{DBLP:conf/atal/AmgoudD19}
Amgoud, L.; and Doder, D. 2019.
\newblock Gradual Semantics Accounting for Varied-Strength Attacks.
\newblock In \emph{Proceedings of the 18th International Conference on
  Autonomous Agents and MultiAgent Systems, {AAMAS} '19, Montreal, QC, Canada},
  1270--1278.

\bibitem[{Amgoud, Doder, and Vesic(2022)}]{DBLP:journals/ai/AmgoudDV22}
Amgoud, L.; Doder, D.; and Vesic, S. 2022.
\newblock Evaluation of argument strength in attack graphs: Foundations and
  semantics.
\newblock \emph{Artificial Intelligence}, 302.

\bibitem[{Amgoud and Prade(2009)}]{DBLP:journals/ai/AmgoudP09}
Amgoud, L.; and Prade, H. 2009.
\newblock Using arguments for making and explaining decisions.
\newblock \emph{Artificial Intelligence}, 173(3-4): 413--436.

\bibitem[{Atkinson et~al.(2017)Atkinson, Baroni, Giacomin, Hunter, Prakken,
  Reed, Simari, Thimm, and Villata}]{Atkinson:17}
Atkinson, K.; Baroni, P.; Giacomin, M.; Hunter, A.; Prakken, H.; Reed, C.;
  Simari, G.~R.; Thimm, M.; and Villata, S. 2017.
\newblock Towards Artificial Argumentation.
\newblock \emph{{AI} Magazine}, 38(3): 25--36.

\bibitem[{Baroni, Caminada, and Giacomin(2011)}]{DBLP:journals/ker/BaroniCG11}
Baroni, P.; Caminada, M.; and Giacomin, M. 2011.
\newblock An introduction to argumentation semantics.
\newblock \emph{Knowl. Eng. Rev.}, 26(4): 365--410.

\bibitem[{Baroni et~al.(2015)Baroni, Romano, Toni, Aurisicchio, and
  Bertanza}]{DBLP:journals/argcom/BaroniRTAB15}
Baroni, P.; Romano, M.; Toni, F.; Aurisicchio, M.; and Bertanza, G. 2015.
\newblock Automatic evaluation of design alternatives with quantitative
  argumentation.
\newblock \emph{Argument Comput.}, 6(1): 24--49.

\bibitem[{Becker and Kohavi(1996)}]{misc_adult_2}
Becker, B.; and Kohavi, R. 1996.
\newblock Adult dataset. UCI Machine Learning Repository.
\newblock https://doi.org/10.24432/C5XW20.

\bibitem[{Besnard and Hunter(2001)}]{DBLP:journals/ai/BesnardH01}
Besnard, P.; and Hunter, A. 2001.
\newblock A logic-based theory of deductive arguments.
\newblock \emph{Artificial Intelligence}, 128(1-2): 203--235.

\bibitem[{Brarda, Tamargo, and
  Garc{\'{\i}}a(2021)}]{DBLP:journals/expert/BrardaTG21}
Brarda, M. E.~B.; Tamargo, L.~H.; and Garc{\'{\i}}a, A.~J. 2021.
\newblock Using Argumentation to Obtain and Explain Results in a Decision
  Support System.
\newblock \emph{{IEEE} Intell. Syst.}, 36(2): 36--42.

\bibitem[{Calders, Kamiran, and Pechenizkiy(2009)}]{DBLP:conf/icdm/CaldersKP09}
Calders, T.; Kamiran, F.; and Pechenizkiy, M. 2009.
\newblock Building Classifiers with Independency Constraints.
\newblock In \emph{{ICDM} Workshops 2009, {IEEE} International Conference on
  Data Mining Workshops, Miami, Florida, USA}, 13--18.

\bibitem[{Chakraborty, Peng, and
  Menzies(2020)}]{DBLP:conf/kbse/ChakrabortyPM20}
Chakraborty, J.; Peng, K.; and Menzies, T. 2020.
\newblock Making Fair {ML} Software using Trustworthy Explanation.
\newblock In \emph{35th {IEEE/ACM} International Conference on Automated
  Software Engineering, {ASE} 2020, Melbourne, Australia}, 1229--1233.

\bibitem[{Cocarascu, Rago, and Toni(2019)}]{DBLP:conf/atal/CocarascuRT19}
Cocarascu, O.; Rago, A.; and Toni, F. 2019.
\newblock Extracting Dialogical Explanations for Review Aggregations with
  Argumentative Dialogical Agents.
\newblock In \emph{Proceedings of the 18th International Conference on
  Autonomous Agents and MultiAgent Systems, {AAMAS} '19, Montreal, QC, Canada},
  1261--1269.

\bibitem[{Cyras et~al.(2019)Cyras, Letsios, Misener, and
  Toni}]{DBLP:conf/aaai/CyrasLMT19}
Cyras, K.; Letsios, D.; Misener, R.; and Toni, F. 2019.
\newblock Argumentation for Explainable Scheduling.
\newblock In \emph{The Thirty-Third {AAAI} Conference on Artificial
  Intelligence, {AAAI} 2019}, 2752--2759.

\bibitem[{Cyras et~al.(2021)Cyras, Rago, Albini, Baroni, and Toni}]{Cyras:21}
Cyras, K.; Rago, A.; Albini, E.; Baroni, P.; and Toni, F. 2021.
\newblock Argumentative {XAI:} {A} Survey.
\newblock In \emph{Proceedings of the Thirtieth International Joint Conference
  on Artificial Intelligence, {IJCAI} 2021, Virtual Event / Montreal, Canada},
  4392--4399.

\bibitem[{da~Costa~Pereira, Tettamanzi, and
  Villata(2011)}]{DBLP:conf/ijcai/PereiraTV11}
da~Costa~Pereira, C.; Tettamanzi, A.; and Villata, S. 2011.
\newblock Changing One's Mind: Erase or Rewind?
\newblock In \emph{{IJCAI} 2011, Proceedings of the 22nd International Joint
  Conference on Artificial Intelligence, Barcelona, Catalonia, Spain},
  164--171.

\bibitem[{Dung(1995)}]{DBLP:journals/ai/Dung95}
Dung, P.~M. 1995.
\newblock On the Acceptability of Arguments and its Fundamental Role in
  Nonmonotonic Reasoning, Logic Programming and n-Person Games.
\newblock \emph{Artificial Intelligence}, 77(2): 321--358.

\bibitem[{Dwork et~al.(2012)Dwork, Hardt, Pitassi, Reingold, and
  Zemel}]{DBLP:conf/innovations/DworkHPRZ12}
Dwork, C.; Hardt, M.; Pitassi, T.; Reingold, O.; and Zemel, R.~S. 2012.
\newblock Fairness through awareness.
\newblock In \emph{Innovations in Theoretical Computer Science 2012, Cambridge,
  MA, USA}, 214--226. {ACM}.

\bibitem[{Efstathiou(2011)}]{DBLP:phd/ethos/Efstathiou11}
Efstathiou, V. 2011.
\newblock \emph{Algorithms for computational argumentation in artificial
  intelligence}.
\newblock Ph.D. thesis, University College London, {UK}.

\bibitem[{E{\u{g}}ilmez, Martins, and Leite(2014)}]{leite-martins:13}
E{\u{g}}ilmez, S.; Martins, J.; and Leite, J. 2014.
\newblock Extending Social Abstract Argumentation with Votes on Attacks.
\newblock In \emph{Theory and Applications of Formal Argumentation}, 16--31.
  Berlin, Heidelberg: Springer Berlin Heidelberg.
\newblock ISBN 978-3-642-54373-9.

\bibitem[{Feldman et~al.(2015)Feldman, Friedler, Moeller, Scheidegger, and
  Venkatasubramanian}]{DBLP:conf/kdd/FeldmanFMSV15}
Feldman, M.; Friedler, S.~A.; Moeller, J.; Scheidegger, C.; and
  Venkatasubramanian, S. 2015.
\newblock Certifying and Removing Disparate Impact.
\newblock In \emph{Proceedings of the 21th {ACM} {SIGKDD} International
  Conference on Knowledge Discovery and Data Mining, Sydney, NSW, Australia},
  259--268.

\bibitem[{Fish, Kun, and Lelkes(2016)}]{DBLP:conf/sdm/FishKL16}
Fish, B.; Kun, J.; and Lelkes, {\'{A}}.~D. 2016.
\newblock A Confidence-Based Approach for Balancing Fairness and Accuracy.
\newblock In \emph{Proceedings of the 2016 {SIAM} International Conference on
  Data Mining, Miami, Florida, USA}, 144--152.

\bibitem[{Fleisher(2021)}]{DBLP:conf/aies/Fleisher21}
Fleisher, W. 2021.
\newblock What's Fair about Individual Fairness?
\newblock In \emph{{AIES} '21: {AAAI/ACM} Conference on AI, Ethics, and
  Society, Virtual Event, USA}, 480--490.

\bibitem[{Gabbay and Rodrigues(2014)}]{gabbay-rodrigues-jlc:13}
Gabbay, D.~M.; and Rodrigues, O. 2014.
\newblock An Equational Approach to the Merging of Argumentation Networks.
\newblock \emph{Journal of Logic and Computation}, 24: 1253--1277.

\bibitem[{Gabbay and Rodrigues(2015)}]{esnan:15}
Gabbay, D.~M.; and Rodrigues, O. 2015.
\newblock Equilibrium States in Numerical Argumentation Networks.
\newblock \emph{Logica Universalis}, 1--63.

\bibitem[{Garg, Villasenor, and Foggo(2020)}]{DBLP:conf/bigdataconf/GargVF20}
Garg, P.; Villasenor, J.~D.; and Foggo, V. 2020.
\newblock Fairness Metrics: {A} Comparative Analysis.
\newblock In \emph{2020 {IEEE} International Conference on Big Data {(IEEE}
  BigData 2020), Atlanta, GA, USA}, 3662--3666.

\bibitem[{Gillingham(2019)}]{gillinghamDecisionSupportSystems2019}
Gillingham, P. 2019.
\newblock Decision support systems, social justice and algorithmic
  accountability in social work: A new challenge.
\newblock \emph{Practice}, 31(4): 277--290.

\bibitem[{Haeri and Zweig(2020)}]{DBLP:conf/ssci/HaeriZ20}
Haeri, M.~A.; and Zweig, K.~A. 2020.
\newblock The Crucial Role of Sensitive Attributes in Fair Classification.
\newblock In \emph{2020 {IEEE} Symposium Series on Computational Intelligence,
  {SSCI} 2020, Canberra, Australia}, 2993--3002.

\bibitem[{Hamon et~al.(2022)Hamon, Junklewitz, Sanchez, Malgieri, and
  de~Hert}]{Hamon:22}
Hamon, R.; Junklewitz, H.; Sanchez, I.; Malgieri, G.; and de~Hert, P. 2022.
\newblock Bridging the Gap Between {AI} and Explainability in the {GDPR:}
  Towards Trustworthiness-by-Design in Automated Decision-Making.
\newblock \emph{{IEEE} Computational Intelligence Magazine}, 17(1): 72--85.

\bibitem[{Hort et~al.(2022)Hort, Chen, Zhang, Sarro, and
  Harman}]{DBLP:journals/corr/abs-2207-07068}
Hort, M.; Chen, Z.; Zhang, J.~M.; Sarro, F.; and Harman, M. 2022.
\newblock Bias Mitigation for Machine Learning Classifiers: {A} Comprehensive
  Survey.
\newblock \emph{CoRR}, abs/2207.07068.

\bibitem[{Hu et~al.(2020)Hu, Iosifidis, Liao, Zhang, Yang, Ntoutsi, and
  Rosenhahn}]{DBLP:conf/dis/HuILZYNR20}
Hu, T.; Iosifidis, V.; Liao, W.; Zhang, H.; Yang, M.~Y.; Ntoutsi, E.; and
  Rosenhahn, B. 2020.
\newblock FairNN - Conjoint Learning of Fair Representations for Fair
  Decisions.
\newblock In \emph{Discovery Science - 23rd International Conference, {DS}
  2020, Thessaloniki, Greece}, volume 12323 of \emph{Lecture Notes in Computer
  Science}, 581--595.

\bibitem[{Iosifidis and Ntoutsi(2019)}]{DBLP:conf/cikm/IosifidisN19}
Iosifidis, V.; and Ntoutsi, E. 2019.
\newblock AdaFair: Cumulative Fairness Adaptive Boosting.
\newblock In \emph{Proceedings of the 28th {ACM} International Conference on
  Information and Knowledge Management, {CIKM} 2019, Beijing, China}, 781--790.

\bibitem[{Kamiran and Calders(2011)}]{DBLP:journals/kais/KamiranC11}
Kamiran, F.; and Calders, T. 2011.
\newblock Data preprocessing techniques for classification without
  discrimination.
\newblock \emph{Knowl. Inf. Syst.}, 33(1): 1--33.

\bibitem[{Kamiran, Calders, and Pechenizkiy(2010)}]{DBLP:conf/icdm/KamiranCP10}
Kamiran, F.; Calders, T.; and Pechenizkiy, M. 2010.
\newblock Discrimination Aware Decision Tree Learning.
\newblock In \emph{{ICDM} 2010, The 10th {IEEE} International Conference on
  Data Mining, Sydney, Australia}, 869--874.

\bibitem[{K{\"{o}}kciyan et~al.(2020)K{\"{o}}kciyan, Parsons, Sassoon, Sklar,
  and Modgil}]{DBLP:conf/eumas/KokciyanPSSM20}
K{\"{o}}kciyan, N.; Parsons, S.; Sassoon, I.; Sklar, E.; and Modgil, S. 2020.
\newblock An Argumentation-Based Approach to Generate Domain-Specific
  Explanations.
\newblock In \emph{Multi-Agent Systems and Agreement Technologies - 17th
  European Conference, {EUMAS} 2020, and 7th International Conference, {AT}
  2020, Thessaloniki, Greece}, volume 12520 of \emph{Lecture Notes in Computer
  Science}, 319--337.

\bibitem[{Larson et~al.(2016)Larson, Mattu, Kirchner, and
  Angwin}]{larsonHowWeAnalyzed2016}
Larson, J.; Mattu, S.; Kirchner, L.; and Angwin, J. 2016.
\newblock How {{We Analyzed}} the {{COMPAS Recidivism Algorithm}}.
\newblock
  https://www.propublica.org/article/how-we-analyzed-the-compas-recidivism-algorithm.

\bibitem[{{Le Quy} et~al.(2022){Le Quy}, Roy, Iosifidis, Zhang, and
  Ntoutsi}]{DBLP:journals/widm/QuyRIZN22}
{Le Quy}, T.; Roy, A.; Iosifidis, V.; Zhang, W.; and Ntoutsi, E. 2022.
\newblock A survey on datasets for fairness-aware machine learning.
\newblock \emph{WIREs Data Mining Knowl. Discov.}, 12(3).

\bibitem[{Leibe, Mikolajczyk, and Schiele(2006)}]{DBLP:conf/bmvc/LeibeMS06}
Leibe, B.; Mikolajczyk, K.; and Schiele, B. 2006.
\newblock Efficient Clustering and Matching for Object Class Recognition.
\newblock In \emph{Proceedings of the British Machine Vision Conference 2006,
  Edinburgh, UK}, 789--798. British Machine Vision Association.

\bibitem[{Leite and Martins(2011)}]{DBLP:conf/ijcai/LeiteM11}
Leite, J.; and Martins, J.~G. 2011.
\newblock Social Abstract Argumentation.
\newblock In \emph{{IJCAI} 2011, Proceedings of the 22nd International Joint
  Conference on Artificial Intelligence, Barcelona, Catalonia, Spain},
  2287--2292.

\bibitem[{Lohia et~al.(2019)Lohia, Ramamurthy, Bhide, Saha, Varshney, and
  Puri}]{DBLP:conf/icassp/LohiaRBSVP19}
Lohia, P.~K.; Ramamurthy, K.~N.; Bhide, M.; Saha, D.; Varshney, K.~R.; and
  Puri, R. 2019.
\newblock Bias Mitigation Post-processing for Individual and Group Fairness.
\newblock In \emph{{IEEE} International Conference on Acoustics, Speech and
  Signal Processing, {ICASSP} 2019, Brighton, UK}, 2847--2851.

\bibitem[{Mehrabi et~al.(2022)Mehrabi, Morstatter, Saxena, Lerman, and
  Galstyan}]{Mehrabi:21}
Mehrabi, N.; Morstatter, F.; Saxena, N.; Lerman, K.; and Galstyan, A. 2022.
\newblock A Survey on Bias and Fairness in Machine Learning.
\newblock \emph{{ACM} Computing Surveys}, 54(6): 115:1--115:35.

\bibitem[{Moro, Rita, and Cortez(2012)}]{misc_bank_marketing_222}
Moro, S.; Rita, P.; and Cortez, P. 2012.
\newblock {Bank Marketing, UCI Machine Learning Repository}.
\newblock https://doi.org/10.24432/C5K306.

\bibitem[{Mukherjee et~al.(2020)Mukherjee, Yurochkin, Banerjee, and
  Sun}]{DBLP:conf/icml/MukherjeeYBS20}
Mukherjee, D.; Yurochkin, M.; Banerjee, M.; and Sun, Y. 2020.
\newblock Two Simple Ways to Learn Individual Fairness Metrics from Data.
\newblock In \emph{Proceedings of the 37th International Conference on Machine
  Learning, {ICML} 2020, 13-18 July 2020, Virtual Event}, volume 119 of
  \emph{Proceedings of Machine Learning Research}, 7097--7107. {PMLR}.

\bibitem[{{Northpointe}(2019)}]{northpointePractitionerGuideCOMPAS2019}
{Northpointe}. 2019.
\newblock Practitioner's {{Guide}} to {{COMPAS Core}}.
\newblock
  https://s3.documentcloud.org/documents/2840784/\\Practitioner-s-Guide-to-COMPAS-Core.pdf.

\bibitem[{Oneto et~al.(2019)Oneto, Donini, Elders, and
  Pontil}]{DBLP:conf/aies/OnetoDEP19}
Oneto, L.; Donini, M.; Elders, A.; and Pontil, M. 2019.
\newblock Taking Advantage of Multitask Learning for Fair Classification.
\newblock In \emph{Proceedings of the 2019 {AAAI/ACM} Conference on AI, Ethics,
  and Society, {AIES} 2019, Honolulu, HI, USA}, 227--237.

\bibitem[{Potyka(2018)}]{DBLP:conf/kr/Potyka18}
Potyka, N. 2018.
\newblock Continuous Dynamical Systems for Weighted Bipolar Argumentation.
\newblock In \emph{Principles of Knowledge Representation and Reasoning:
  Proceedings of the Sixteenth International Conference, {KR} 2018, Tempe,
  Arizona}, 148--157.

\bibitem[{Pu et~al.(2014)Pu, Luo, Zhang, and Luo}]{DBLP:conf/ksem/PuLZL14}
Pu, F.; Luo, J.; Zhang, Y.; and Luo, G. 2014.
\newblock Argument Ranking with Categoriser Function.
\newblock In \emph{Knowledge Science, Engineering and Management - 7th
  International Conference, {KSEM} 2014, Sibiu, Romania}, volume 8793 of
  \emph{Lecture Notes in Computer Science}, 290--301.

\bibitem[{Rago, Cocarascu, and Toni(2018)}]{DBLP:conf/ijcai/RagoCT18}
Rago, A.; Cocarascu, O.; and Toni, F. 2018.
\newblock Argumentation-Based Recommendations: Fantastic Explanations and How
  to Find Them.
\newblock In \emph{Proceedings of the Twenty-Seventh International Joint
  Conference on Artificial Intelligence, {IJCAI} 2018, Stockholm, Sweden},
  1949--1955.

\bibitem[{{The Partnership on
  AI}(2019)}]{thepartnershiponaiReportAlgorithmicRisk2019}
{The Partnership on AI}. 2019.
\newblock Report on {{Algorithmic Risk Assessment Tools}} in the {{U}}.{{S}}.
  {{Criminal Justice System}}.
\newblock
  https://www.partnershiponai.org/report-on-machine-learning-in-risk-assessment-tools-in-the-u-s-criminal-justice-system/.

\bibitem[{Tilmes(2022)}]{DBLP:journals/ethicsit/Tilmes22}
Tilmes, N. 2022.
\newblock Disability, fairness, and algorithmic bias in {AI} recruitment.
\newblock \emph{Ethics Inf. Technol.}, 24(2): 21.

\bibitem[{Vassiliades, Bassiliades, and Patkos(2021)}]{Vassiliades:P21}
Vassiliades, A.; Bassiliades, N.; and Patkos, T. 2021.
\newblock Argumentation and explainable artificial intelligence: a survey.
\newblock \emph{Knowledge Engineering Review}, 36: e5.

\bibitem[{Verma and Rubin(2018)}]{Verma:18}
Verma, S.; and Rubin, J. 2018.
\newblock Fairness definitions explained.
\newblock In \emph{Proceedings of the International Workshop on Software
  Fairness, FairWare@ICSE 2018, Gothenburg, Sweden}, 1--7.

\bibitem[{Wachter, Mittelstadt, and
  Russell(2021)}]{DBLP:journals/clsr/WachterMR21}
Wachter, S.; Mittelstadt, B.~D.; and Russell, C. 2021.
\newblock Why fairness cannot be automated: Bridging the gap between {EU}
  non-discrimination law and {AI}.
\newblock \emph{Comput. Law Secur. Rev.}, 41: 105567.

\bibitem[{Waller, Rodrigues, and Cocarascu(2023)}]{waller2023bias}
Waller, M.; Rodrigues, O.; and Cocarascu, O. 2023.
\newblock Bias Mitigation Methods for Binary Classification Decision-Making
  Systems: Survey and Recommendations.
\newblock \emph{CoRR}, abs/2305.20020.

\bibitem[{Weinberg(2022)}]{DBLP:journals/jair/Weinberg22}
Weinberg, L. 2022.
\newblock Rethinking Fairness: An Interdisciplinary Survey of Critiques of
  Hegemonic {ML} Fairness Approaches.
\newblock \emph{Journal of Artificial Intelligence Research (JAIR)}, 74:
  75--109.

\bibitem[{Zemel et~al.(2013)Zemel, Wu, Swersky, Pitassi, and
  Dwork}]{DBLP:conf/icml/ZemelWSPD13}
Zemel, R.~S.; Wu, Y.; Swersky, K.; Pitassi, T.; and Dwork, C. 2013.
\newblock Learning Fair Representations.
\newblock In \emph{Proceedings of the 30th International Conference on Machine
  Learning, {ICML} 2013, Atlanta, GA, USA3}, volume~28 of \emph{{JMLR} Workshop
  and Conference Proceedings}, 325--333.

\bibitem[{Zliobaite, Kamiran, and Calders(2011)}]{DBLP:conf/icdm/ZliobaiteKC11}
Zliobaite, I.; Kamiran, F.; and Calders, T. 2011.
\newblock Handling Conditional Discrimination.
\newblock In \emph{11th {IEEE} International Conference on Data Mining, {ICDM}
  2011, Vancouver, BC, Canada}, 992--1001.

\end{thebibliography}

\end{document}